\documentclass{article}
\usepackage{spconf,amsmath,epsfig}
\usepackage{tabularx}
\usepackage{graphicx}
\usepackage{multirow}
\usepackage{multicol}
\usepackage{floatrow}
\usepackage{subfigure}
\usepackage{makecell,rotating}
\usepackage{textcomp,booktabs}
\usepackage[usenames,dvipsnames]{color}
\usepackage{colortbl}
\definecolor{setgray}{gray}{.95}
\definecolor{setblue}{gray}{.7}
\newcolumntype{C}[1]{>{\centering\arraybackslash}p{#1}}
\newcolumntype{L}[1]{>{\raggedright\arraybackslash}p{#1}}
\usepackage{amssymb}
\usepackage{amsmath}
\usepackage{bm}
\usepackage{cite}
\usepackage[colorlinks=true,linkcolor=black,citecolor=black]{hyperref}
\usepackage{url}

\title{Large-scale Land Cover Classification in GaoFen-2 Satellite Imagery
\thanks{This work is supported by NSFC projects under the contracts No.61771350 and No.41501462.}}

\name{Xin-Yi Tong${}^1$ , Qikai Lu${}^2$, Gui-Song Xia${}^1$, Liangpei Zhang${}^1$}
\address{${}^1$ State Key Laboratory of LIESMARS, Wuhan University, Wuhan 430079, China\\
${}^2$ Electronic Information School, Wuhan University, Wuhan 430072, China}

\begin{document}

\maketitle

\begin{abstract}
Many significant applications need land cover information of remote sensing images that are acquired from different areas and times, such as change detection and disaster monitoring. However, it is difficult to find a generic land cover classification scheme for different remote sensing images due to the spectral shift caused by diverse acquisition condition. In this paper, we develop a novel land cover classification method that can deal with large-scale data captured from widely distributed areas and different times. Additionally, we establish a large-scale land cover classification dataset consisting of 150 Gaofen-2 imageries as data support for model training and performance evaluation. Our experiments achieve outstanding classification accuracy compared with traditional methods.

\end{abstract}

\begin{keywords}
land cover classification, high-resolution remote sensing image, Gaofen-2, deep learning
\end{keywords}
\section{Introduction}
\label{sec:intro}
Land cover classification plays a crucial role in applications such as land resource management, urban planning and environmental protection \cite{application}. With the development of remote sensing technology, imaging satellites can supply remote sensing data that cover most of the Earth¡¯s surface, which provides new chances to land cover classification, while also bringing great challenges. In images captured from different areas and at different times, the photographic distortion, viewing angle, scale, illumination and observed content vary largely, which makes it difficult to find an efficient land cover classification method for different remote sensing images.

Up to now, land cover classification task has been widely investigated. Primary studies interpret the image information according to the spectrum of every individual pixel. However, this kind of methods is easy to be affected by intra-class spectral variability and spectral noise \cite{spectrum2}. More recent methods implement spectral-spatial classification, which integrates regional spatial information to boost the performance of classification. Spectral-spatial features, such as texture, shape and structure, can represent more characteristics of the images than pure spectral or spectral derivative features \cite{spatialspectral2}. However, neither spectrum nor spectral-spatial features have sufficient invariance to complex changes emerging in various RS data. When processing new images, new labeled data and algorithm adjustment are generally necessary, which reduces the efficiency of practical applications.

In recent years, as deep learning has achieved breakthrough in the field of visual recognition, new path has been set for image categorization in the remote sensing community \cite{DLreview}. Convolutional Neural Networks (CNN) are the most representative deep learning models, which are constructed in deep hierarchical architectures and capable of extracting the intrinsic features of data \cite{RSDL,AID}. Researchers have utilized deep features to promote the performance of land cover classification \cite{dl1,dl2}. Nevertheless, the recognition capacity of CNN models greatly depends on the size of annotated training data. The existing high-resolution land cover datasets either cover limited geographic areas with insufficient samples \cite{data1} or cover homogeneous areas with low intra-class variability \cite{data2} . These limitations incline to restrain the generalization ability of deep models.

In this paper, in order to efficiently classify remote sensing images captured under different conditions, we propose a novel method based on the combination of deep learning, hierarchical segmentation and multi-scale information fusion. We establish a large-scale land cover dataset with 150 Gaofen-2 (GF-2) imageries to support our approach. This dataset has high intra-class differences and low inter-class diversities and hence it can also serve as data resource to advance the state-of-the-art in land cover classification task. Our experiments achieve outstanding classification performance compared with traditional methods.

\begin{figure}
\centering
\includegraphics[width=1\textwidth]
{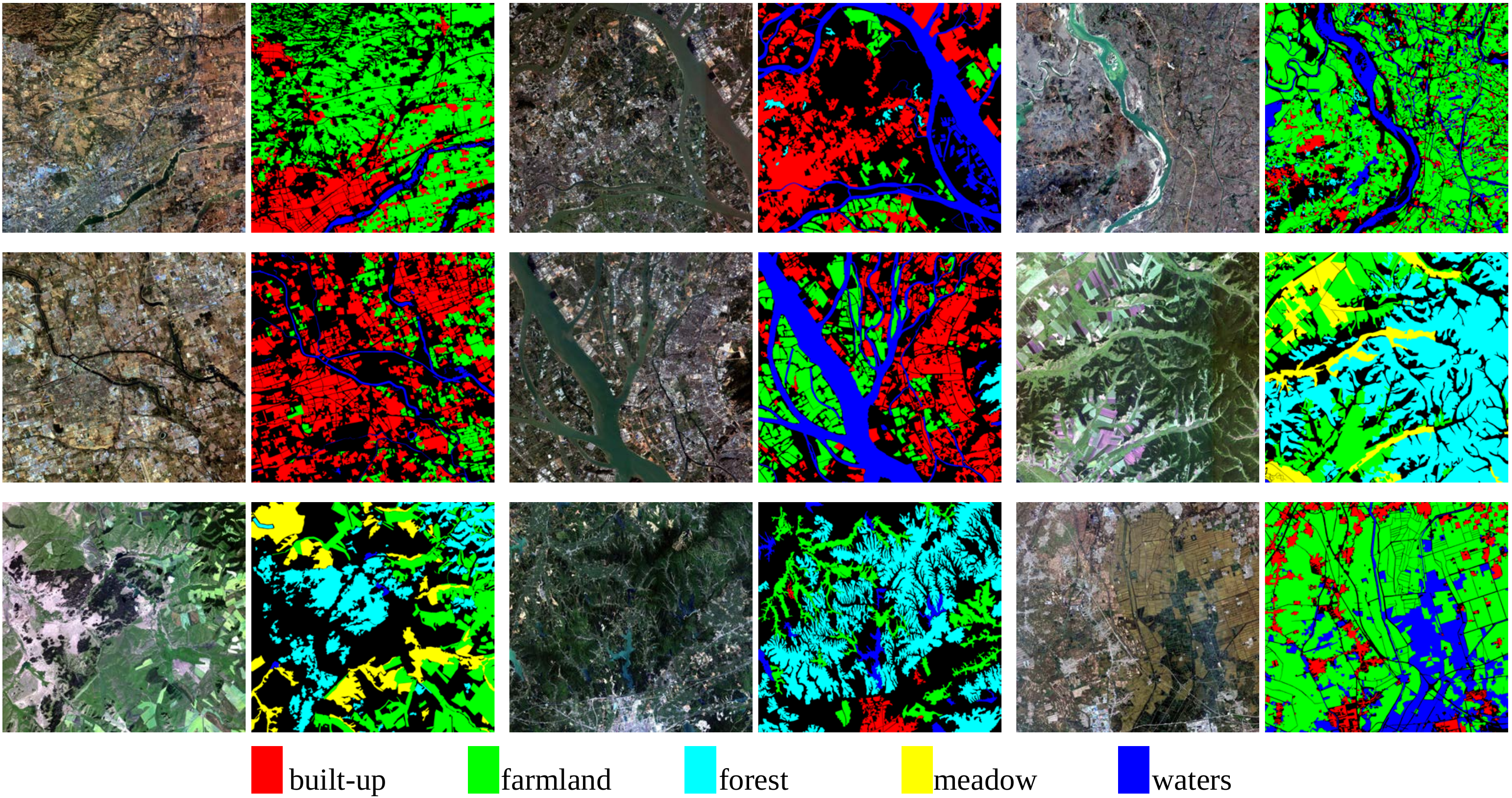}
\caption{Some samples and corresponding label masks of GID.}
\label{figure:sample}
\end{figure}

\section{Dataset Description}
\label{sec:data}

\subsection{Gaofen-2 Satellite Imagery}
GF-2 satellite is configured with two Panchromatic and Multispectral CCD Camera Sensors (PMS), which have a resolution of 1 m panchromatic/4 m multispectral. It can provide a combined swath of 45 km, which is embodied as a size of $6908\times7300$ pixels in multispectral imagery. The revisiting period of GF-2 is 5 days, hence it is able to capture detail information over a wide area at short intervals. Consider the combination of high resolution, wide imaging coverage, frequent revisit and high image quality, GF-2 imagery is an ideal data source for land cover classification.
\subsection{Study Area}
We annotate 150 GF-2 satellite images to construct a large-scale land cover dataset, which is named as Gaofen Image Dataset (GID). GID is widely distributed over the geographic areas covering more than 70,000 km$^{2}$. Benefit from the various acquisition locations and times, GID presents rich diversity in spectral response and morphological structure. Five representative land cover categories of application values are selected to be annotated: built-up, farmland, forest, meadow, and waters. Areas that do not belong to the above five categories or cannot be artificially recognized are labeled as unknown, which is represented using black color. Fig. \ref{figure:sample} shows some samples and their corresponding label masks of GID.

\section{Methodology}
\label{sec:method}
\begin{figure*}
\centering
\includegraphics[width=1\textwidth]
{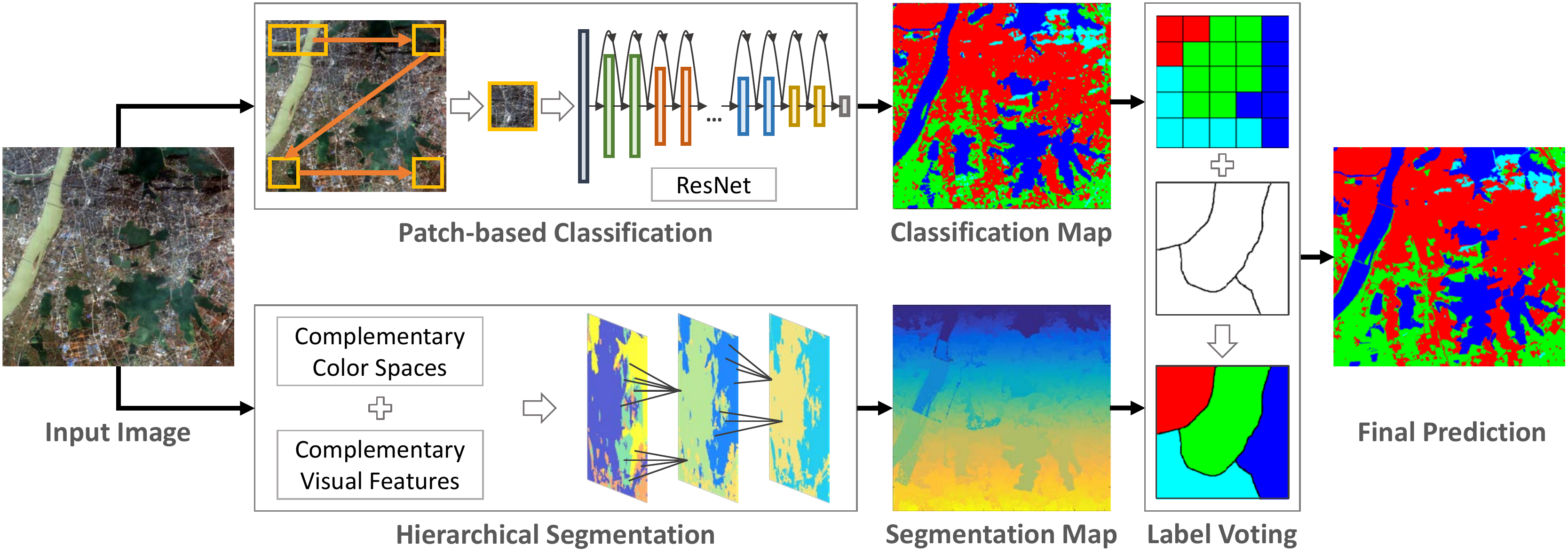}
\caption{The flow chart of our method.}
\label{figure:framwork}
\end{figure*}

In this section, we describe the proposed land cover classification algorithm in detail. Our method combines deep learning, hierarchical segmentation and multi-scale information fusion. Specifically, we firstly use convolutional neural networks (CNN) to classify GF-2 imageries in form of image patches, and simultaneously use segmentation method to obtain a series of homogeneous objects. Then we fuse the classification and the segmentation maps with voting strategy, which is illustrated in Fig. \ref{figure:framwork}. Finally, multi-scale spatial information is collected to augment the context information and further promote the classification performance.

\subsection{Patch-based CNN classification}
Residual networks (ResNet) \cite{resnet} has many advantages over the previous CNN models. It relieves the problem of the gradient disappearance and is easier to train when the net architecture is very deep. We re-train ResNet-50 model with GID to conduct patch-based classification. It has a total of 49 convolution layers, consisting 16 bottleneck structures. Each of these bottleneck structures has three convolutional layers, which first decrease and then elevate the dimension of feature maps to control the number of parameters.

When fine-tuning, we cut the original GF-2 imagery into compact square patches, and then randomly select patches as training samples. We remove the 1000-dimensional softmax layer of ResNet-50 and change it into a Gaussian distribution initialized 5-dimensional softmax layer, where 5 is the number of land cover categories in GID. We only fine-tune ResNet-50's last three bottleneck structures and the last softmax layer. The hyper-parameters are set as follows: batch size is 32, epoch is 15, momentum is 0.9, and initial learning rate is set to be 0.1. During the iteration, when the error rate stops decreasing, we divide the learning rate by 10 and use this new value to update the parameters. In the experiment, the learning rate was reduced three times before the model is converged.

In the process of classification, we cut the testing imageries into square patches with the same size as the training samples, and then acquire their category distribution probability from re-trained ResNet-50's softmax layer. The entire GF-2 imagery is thus classified in the form of compact patches.

\subsection{Segmentation and Voting}
Due to the input limitation of CNN¡¯s structure, we classify GF-2 imageries on the basis of image patches, which completely loses the boundary information of ground objects. Considering the above issue, we utilize selective search \cite{segmentation} segmentation as post-processing. Selective search segmentation employs a graph-based approach to obtain a variety of initial regions in different color spaces, and then iteratively merge the small regions into bigger ones with greedy algorithm. There are different consolidation strategies controlling the level of merging, so this method can accurately extract the object information in remote sensing images.

After obtaining the results of classification and segmentation, we use voting strategy to combine category and boundary information. For every segmented region, the number of pixels belonging to different categories is counted. Each pixel belonging to a same segmented region votes for its corresponding class, and the entire region is labeled with the category that gets the most votes. After completely voting for each segmented region in the whole GF-2 imagery, we obtain the final classification map.

\subsection{Multi-scale Classification}
Despite that CNN models have certain invariance to rotation, translation, and illumination change, their fixed input size and receptive fields limit their observable space. When identifying the category of an area in remote sensing image, not only the local information but also the spatial context information is crucial. Therefore, we fuse multi-scale spatial information extracted from the same locations to acquire the context information.

In consideration of the computational efficiency and the parameter size of deep model, we train a single model with multi-scale patches. We randomly select a certain number of sampling points from GF-2 imageries for each category and treat these points as square centers to cut image patches with different scales respectively. We warp them into a uniform size and use them to fine-tune ResNet-50 model. When testing, we sum up the classification probability vectors of different scales and then use the summed probability vector to identify land cover category.

\begin{table*}[!tbp]
\caption{Comparison of training samples of different scales.}
\arrayrulecolor{setblue}
\renewcommand\arraystretch{1.1}
\resizebox{\textwidth}{!}{
\begin{tabular}{ccccccccccc}
\toprule[1pt]
\rowcolor{setgray}
Scales&&\textbf{Kappa}&\textbf{OA}&\textbf{AA}&&built-up&farmland&forest&meadow&waters\\
\midrule[0.8pt]
$56\times56$ pixels&&0.8958&94.94\%&\textbf{88.35\%}&&92.37\%&94.91\%&87.98\%&92.27\%&82.49\%\\
\cmidrule[0.8pt]{1-1}
$112\times112$ pixels&&0.8731&93.79\%&83.00\%&&86.98\%&92.52\%&74.98\%&75.89\%&82.18\%\\
\cmidrule[0.8pt]{1-1}
$224\times224$ pixels&&0.8350&91.85\%&78.27\%&&80.54\%&87.44\%&70.46\%&90.85\%&79.00\%\\
\cmidrule[0.8pt]{1-1}
multi-scale&&\textbf{0.9129}&\textbf{95.71\%}&87.12\%&&87.97\%&95.60\%&84.52\%&78.64\%&88.41\%\\
\bottomrule[1pt]
\end{tabular}}
\label{table:I}
\end{table*}

\begin{table*}[!tbp]
\caption{Comparison of different land cover classification methods.}
\arrayrulecolor{setblue}
\renewcommand\arraystretch{1.1}
\resizebox{\textwidth}{!}{
\begin{tabular}{ccccccccccc}
\toprule[1pt]
\rowcolor{setgray}
Methods&&\textbf{Kappa}&\textbf{OA}&\textbf{AA}&&built-up&farmland&forest&meadow&waters\\
\midrule[0.8pt]
eCognition&&0.5639&74.87\%&69.35\%&&76.03\%&66.89\%&85.83\%&85.39\%&75.34\%\\
\cmidrule[0.8pt]{1-1}
CH+SVM&&0.6072&77.69\%&70.89\%&&78.84\%&76.97\%&56.31\%&67.67\%&65.35\%\\
\cmidrule[0.8pt]{1-1}
GLCM+SVM&&0.5494&72.81\%&69.49\%&&86.61\%&68.89\%&42.71\%&68.47\%&65.09\%\\
\cmidrule[0.8pt]{1-1}
LBP+SVM&&0.4745&67.87\%&64.81\%&&59.06\%&59.24\%&59.18\%&59.62\%&81.83\%\\
\cmidrule[0.8pt]{1-1}
Fusion+SVM&&0.6038&77.28\%&74.19\%&&70.15\%&75.35\%&69.66\%&87.00\%&80.05\%\\
\cmidrule[0.8pt]{1-1}
CH+RF&&0.6340&79.83\%&69.77\%&&88.74\%&80.46\%&31.77\%&42.54\%&62.37\%\\
\cmidrule[0.8pt]{1-1}
GLCM+RF&&0.5964&75.43\%&70.94\%&&89.13\%&75.88\%&51.94\%&43.90\%&57.18\%\\
\cmidrule[0.8pt]{1-1}
LBP+RF&&0.6660&81.95\%&76.64\%&&75.57\%&78.37\%&71.21\%&68.77\%&80.62\%\\
\cmidrule[0.8pt]{1-1}
Fusion+RF&&0.7280&84.75\%&77.44\%&&89.16\%&84.16\%&48.11\%&71.01\%&77.16\%\\
\midrule[0.8pt]
multi-scale (our)&&\textbf{0.9129}&\textbf{95.71\%}&\textbf{87.12}\%&&87.97\%&95.60\%&84.52\%&78.64\%&88.41\%\\
\bottomrule[1pt]
\end{tabular}}
\label{table:II}
\end{table*}

\section{Experiments}
\label{sec:experiment}
\subsection{Experiment settings}
In GID, we randomly select 120 GF-2 imageries as training data and treat the rest 30 imageries as testing data. 30,000 sampling points are randomly selected for every category, constituting a training set of 150,000 samples at the total. We separately set the patch size as $56\times56$, $112\times112$, $224\times224$ pixels. With the exception of $224\times224$-pixel patches, the other sizes are fixed into $224\times224$ pixels before being put into ResNet. In pre-processing, we remove the near-infrared band from GF-2 images, and then re-quantize the response of the visible light band to 8-bit.

For performance comparison, we tested some traditional land cover classification methods. The examined features include color histogram (CH), gray-level co-occurrence matrix (GLCM), local binary patterns (LBP) and their fused features, the classifiers exploited include support vector machine (SVM) and random forest (RF). The method of feature fusion is normalized vector concatenation. In addition, we compare our performance with eCognition software. For traditional methods, the training and testing sets are sampled from a single image. We use overall accuracy (OA), average accuracy (AA) and kappa coefficient to quantitatively evaluate the experimental results.

\subsection{Classification Results}
Table. \ref{table:I} illustrates the resulting kappa, OA, AA and each category accuracy of different sampling scales. For single-scale, it is obvious that the patches with the smallest scale generate the highest accuracy. However, their accuracies are generally lower than those of multi-scale. This is because multi-scale patches incorporate spatial context information from wider observable areas. Table. \ref{table:II} compares the quantitative results of our method and other baseline methods. The accuracy of our method is significantly higher than that of the traditional methods.

\section{Conclusion}
\label{sec:conclusion}
In order to solve the problem of adaptability limitation of LULC classification methods to RS images captured under different conditions, we propose a classification method based on deep learning, hierarchical segmentation and multi-scale information fusion. Meanwhile, we introduce a large-scale land cover classification dataset consisting of 150 Gaofen-2 imageries labeled in 5 categories for model training and performance evaluation. It covers a large area and has high intra-class diversity, hence it can also provide the research community with a high-quality data resource for evaluating and advancing the state-of-the-art methods in land cover classification. The experiments show that, the proposed method can achieve remarkable performance, and the combination of spatial context information of different scales can make the contribution to classification accuracy.

\footnotesize
\bibliographystyle{IEEEbib}
\bibliography{reference}

\end{document}